\documentclass{article}

\PassOptionsToPackage{numbers, compress}{natbib}
\usepackage[final]{neurips_2021}
\usepackage{todonotes}
\usepackage[utf8]{inputenc} % allow utf-8 input
\usepackage[T1]{fontenc}    % use 8-bit T1 fonts
\usepackage{hyperref}       % hyperlinks
\usepackage{url}            % simple URL typesetting
\usepackage{booktabs}       % professional-quality tables
\usepackage{amsfonts}       % blackboard math symbols
\usepackage{nicefrac}       % compact symbols for 1/2, etc.
\usepackage{microtype}      % microtypography
\usepackage{xcolor}         % colors
\usepackage{amsmath} % for softmax function to work
\usepackage{boldline} % used for bold vertical lines in tables
\usepackage{float}
\restylefloat{table}
\usepackage{ctable}
\usepackage{tabularx}

\usepackage{graphicx}
\usepackage{listings}
\usepackage{xcolor}

\colorlet{punct}{red!60!black}
\definecolor{background}{HTML}{EEEEEE}
\definecolor{delim}{RGB}{20,105,176}
\colorlet{numb}{magenta!60!black}

\lstdefinelanguage{json}{
    basicstyle=\small\ttfamily,
    stepnumber=1,
    numbersep=1pt,
    showstringspaces=false,
    breaklines=true,
    frame=lines,
    literate=
     *{0}{{{\color{numb}0}}}{1}
      {1}{{{\color{numb}1}}}{1}
      {2}{{{\color{numb}2}}}{1}
      {3}{{{\color{numb}3}}}{1}
      {4}{{{\color{numb}4}}}{1}
      {5}{{{\color{numb}5}}}{1}
      {6}{{{\color{numb}6}}}{1}
      {7}{{{\color{numb}7}}}{1}
      {8}{{{\color{numb}8}}}{1}
      {9}{{{\color{numb}9}}}{1}
      {:}{{{\color{punct}{:}}}}{1}
      {,}{{{\color{punct}{,}}}}{1}
      {\{}{{{\color{delim}{\{}}}}{1}
      {\}}{{{\color{delim}{\}}}}}{1}
      {[}{{{\color{delim}{[}}}}{1}
      {]}{{{\color{delim}{]}}}}{1},
}

\title{DP-KB: Data Programming with Knowledge Bases Improves Transformer Fine Tuning for Answer Sentence Selection}

\author{
Nic Jedema \\ 
Alexa AI - Graphiq \\
Santa Barbara, CA 93101 \\
\texttt{jedem@amazon.com} \\
\And
Thuy Vu \\ 
Alexa AI - Search \\
Manhattan Beach, CA 90266 \\
\texttt{thuyvu@amazon.com} \\
\And
Manish Gupta \\ 
Alexa AI - Graphiq \\
Santa Barbara, CA 93101 \\
\texttt{manishg@amazon.com} \\
\And
Alessandro Moschitti \\ 
Alexa AI - Search \\
Manhattan Beach, CA 90266 \\
\texttt{amosch@amazon.com} \\
}

\begin{document}

\maketitle

\begin{abstract}
While transformers demonstrate impressive performance on many knowledge intensive (KI) tasks, their ability to serve as implicit knowledge bases (KBs) remains limited, as shown on several slot-filling, 
 question-answering (QA), fact verification, and entity-linking tasks. 
In this paper, we implement an efficient, data-programming technique that enriches training data with KB-derived context and improves transformer utilization of encoded knowledge when fine-tuning for a particular QA task, namely answer sentence selection (AS2).
Our method outperforms state of the art transformer approach on WikiQA and TrecQA, two widely studied AS2 benchmarks, increasing by 2.0\% p@1, 1.3\% MAP, 1.1\% MRR, and 4.4\% p@1, 0.9\% MAP,  2.4\% MRR, respectively. 
To demonstrate our improvements in an industry setting, we additionally evaluate our approach on a proprietary dataset of Alexa QA pairs, and show increase of 2.3\% F1 and 2.0\% MAP.
We additionally find that these improvements remain even when KB context is omitted at inference time, allowing for the use of our models within existing transformer workflows without additional latency or deployment costs. 
\end{abstract}

\section{Introduction}
% 1) Lots of evidence bc transformers act as impl KBs and thus can be used for downstream tasks in closed book/KI tasks
Transformers are powerful sequence-to-sequence models that leverage the mechanism of self-attention  \cite{vaswani2017attention} to capture long-range dependencies, such as relationships between words in a natural language text. Compared to sequential models such as recurrent or convolutional networks, they make efficient use of available processing power. Transformer based models such as BERT, XLNet and BART \cite{devlin2019bert, yang2020xlnet, liu2019roberta, lewis2019bart} hold state of the art on many natural language processing (NLP) tasks, including next sentence prediction, natural language generation, and natural language inference \cite{raffel2020exploring, lewis2019bart, he2021deberta}. In addition to efficiently encoding linguistic information from unlabelled text, their top performance on knowledge-intensive (KI) NLP tasks \cite{petroni2021kilt}, such as question-answering \cite{rajpurkar2016squad} have led to the hypothesis that transformers also encode relational knowledge, and as such serve as parameterized, implicit knowledge bases (KBs) \cite{petroni2019language}.

However, it has also been shown that transformer knowledge acquisition \cite{petroni2019language, roberts2020knowledge} and subsequent utilization \cite{talmor2020olmpics, jiang2020know} can be uncontrollable, highly context dependent, and tightly coupled to language acquisition. 
These limitations may impact performance on downstream tasks, including KI tasks like answer sentence selection (AS2) \cite{wang2007trec}.
Table \ref{table:hard-qa-pairs} is illustrative of limitations of some of these deficiencies of transformers in precisely leveraging encoded information. Transformer models that show top performance \cite{garg2019tanda, laskar-etal-2020-contextualized} on widely studied benchmarks \cite{yang-etal-2015-wikiqa, wang2007trec} still fail to classify many QA pairs correctly.  In \emph{Example 1}, the model is unable to leverage knowledge of the identity between \emph{Elton John's husband} and \emph{David Furnish}. 
In \emph{Example 2}, \emph{one-humped} or \emph{two-humped} are not recognizable as quantities pertaining to the uncommonly quantity \emph{humps}.
\emph{Example 3} shows the difficulty in reasoning for the a rare prescriptive use of the illicit drug \emph{methamphetamine}. 
These examples also illustrate relevance of this task as a means to assess impact of deficiencies in transformer knowledge utilization.

In this paper study, we propose an efficient, data-programming approach utilizing a KB that improves performance on answer selection tasks and demonstrate that some of these limitations can be mitigated during fine-tuning with simple data augmentation technique.

\begin{table}
\caption{Three QA examples incorrectly predicted by a state-of-the-art transformer answer selection model (TANDA \cite{garg2019tanda}).}
\label{table:hard-qa-pairs}
\centering
\small
\begin{tabularx}{\textwidth}{lX}
\hline

 \textit{Example 1}:  &  \emph{\textbf{Q:} How old is Elton John's husband?\newline 
\textbf{Correct:}  David Furnish is 57 years old. He was born on October 25, 1962.\newline 
\textbf{Selected:} Elton John and David Furnish became an item after meeting in the early 1990s and in 2005.
}
\\
\hline
 \textit{Example 2}:   &   \emph{\textbf{Q:} How many humps on a Camel? \newline
 \textbf{Correct:} The two surviving species of camel are the dromedary, or one-humped camel, which is native to the Middle East and the Horn of Africa; and the Bactrian, or two-humped camel, which inhabits Central Asia.\newline
 \textbf{Selected:} A camel is an even-toed ungulate within the genus Camelus, bearing distinctive fatty deposits known as "humps" on its back.
 } 
  \\
 \hline
\textit{Example 3}:   &    \emph{\textbf{Q:} What some legal uses of meth? \newline
\textbf{Correct:} Although rarely prescribed, methamphetamine hydrochloride is approved by the U.S. Food and Drug Administration (FDA) for the treatment of attention deficit hyperactivity disorder and obesity under the trade name Desoxyn.\newline
\textbf{Selected:} Methamphetamine, also known as metamfetamine, meth, ice, crystal, glass, tik, N-methylamphetamine, methylamphetamine, and desoxyephedrine, is a psychostimulant of the phenethylamine and amphetamine class of psychoactive drugs.
} 
\\
\hline
\end{tabularx}
\end{table}

% 3) There are a number of approaches that have been proposed to increase acquisition of knowledge for KI tasks - KnowBERT, ERNIE, KELM. For AS2, context-based approaches have been proposed that treat AS2 as open-book. However, the study of increasing the transformer recall of encoded knowledge in a closed book setting - such as AS2 - has not been studied. 
A number of recent studies have also studied approaches that aim to improve transformer performance on KI tasks, proposing the use of differentiable knowledge retrievers \cite{guu2020realm, karpukhin2020dense, lee2019latent}, retrieval-augmented generation (RAG) \cite{karpukhin2020dense}, KB embeddings such as KnowBERT \cite{peters2019knowledge} and ERNIE \cite{zhang2019ernie}, and pre-training on verbalized KBs such as KELM \cite{agarwal2021knowledge}.
While these approaches offer promising benefits for transformer knowledge encoding and retrieval, to our knowledge, none of them have been shown to outperform existing state of the art for answer selection, a task that is essential to several question answering services provided by commercial voice assistants. 
Additionally, each of these approaches is significantly complex and require significant work to leverage in production applications.
Our approach, on the other hand, leverages ElasticSearch to tag KB entries in input QA pairs, derives weak-supervision signals from tagged KB entries, and incorporates this context only during fine-tuning.
We show that our simple, efficient and data-programming method confers significant performance benefits over the state of the art for answer sentence selection, even when KB context is omitted at inference time.

The main contributions of our work are:
\begin{itemize}
	\item We show that several limitations in the use of transformers implicit KBs can be overcome using a simple data-programming approach by outperforming state-of-the-art models on several QA tasks:
	\begin{enumerate}
	\item increasing by 2.0\% p@1, 1.3\% MAP,  1.1\% MRR  and 4.4\% p@1, 0.9\% MAP,  2.4\% MRR  on WikiQA and TrecQA respectively, two widely used AS2 benchmarks.
	\item increasing by 2.3\% F1 and 2.0\% MAP on AlexaQA pairs, a proprietary commercial answer classification benchmark.
	\end{enumerate}
	\item We show that KB is not needed at inference time, allowing our trained models to be used as drop-in replacements for existing transformer-based AS2 systems.
\end{itemize}

\section{Background}
\subsection{Transformers}
\label{sec:transformers}
The transformer \cite{vaswani2017attention} is an architecture for efficiently transforming one sequence into another via \emph{self-attention}, a mechanism that differentially weighs the significance of discrete  tokens in an input sequence. 
Compared to sequentially aligned or convolutional networks such as RNNs and CNNs, transformer models have proven to be extremely effective at efficiently capturing long-range dependencies between words in natural language \cite{devlin2019bert, liu2019roberta, lewis2019bart}, including some structured knowledge such as the \emph{husband of} relation between named entities \cite{petroni2019language}. 
Pre-training transformer models on large, unstructured corpora of unlabelled text \cite{dai2015semisupervised, howard2018universal} allows them to capture linguistic and factual knowledge prior to subsequent fine-tuning on downstream tasks. 
The suitability of transformer based models such as BERT \cite{devlin2019bert} for this type of transfer learning dramatically increases their reusability, driving state of the art results on many tasks \cite{garg2019tanda, laskar-etal-2020-contextualized, guu2020realm, yang2020xlnet, hardalov2020enriched} in addition to widespread adoption in industry.

\subsection{Transformer Limitations as Knowledge Bases}
\label{sec:transformer-limitations}
While transformers have demonstrated strong performance on question-answering \cite{roberts2020knowledge} and fill-in-the-blank cloze tasks \cite{petroni2019language, jiang2020know} without access to external information, the modulation of transformer knowledge acquisition and utilization is limited.
Cloze task \cite{petroni2019language} and question answering \cite{roberts2020knowledge} probes demonstrate transformer knowledge acquisition is largely uncontrollable and often only results in the acquisition of frequently observed information.
Further, transformer recall of factual knowledge on cloze tasks remains tightly bound to learned linguistic representation \cite{jiang2020know}. 
In a systematic study \cite{talmor2020olmpics} on multiple tasks, it was shown that transformers lack robust multi-hop reasoning faculties, are insensitive to adverbial modifies like "always", "some", and "never", and are unable to robustly compare quantities.
For example, while RoBERTa \cite{liu2019roberta} appears able to effectively compare numbers, it is unable to compare when values are given in \emph{ages}. 
Other studies additionally have shown insensitivity to negation \cite{ettinger-2020-bert}, difficulty with misspellings and short, simple sequences \cite{sun2020advbert}, and sensitivity to sequence length, punctuation, and subject-verb agreement \cite{chernyavskiy2021transformers}.

\subsection{Answer Sentence Selection (AS2) and Answer Classification}
\label{sec:AS2}
Answer sentence selection (AS2) consists of ranking answer candidates given a question and one or more answer candidates, while binary answer classification consists of classifying answer candidates as \emph{correct} or \emph{incorrect} given the same input.
Both tasks encourage models that leverage encoded knowledge to select the most correct answer and thus may be used to probe model knowledge and reasoning capabilities.
Let $q$ be a question, $C_q=\{c_1, \dots, c_n\}$ be a set of answer sentence candidates for $q$, we define $\mathcal{R}$ as a ranking function, which orders the candidates in $C_q$ according to a score, $p\left(q, c_i\right)$, indicating the probability of $c_i$ to be a correct answer for $q$.
Answer sentence selection is performed by taking the highest scoring candidate in $C_q$, while binary answer classification is performed by assigning the label of the highest probability class as determined by the ranking function $\mathcal{R}$ .

Widely used metrics for AS2 performance are \href{https://en.wikipedia.org/wiki/Evaluation_measures_(information_retrieval)#Mean_average_precision}{mean average precision} (MAP) and \href{https://en.wikipedia.org/wiki/Mean_reciprocal_rank}{mean reciprocal rank} (MRR), while \href{https://en.wikipedia.org/wiki/Evaluation_measures_(information_retrieval)#Mean_average_precision}{mean average precision} (MAP) and \href{https://en.wikipedia.org/wiki/F-score}{F-score} (F1) are commonly used for binary answer classification.
To our knowledge, transformer models \cite{garg2019tanda, laskar-etal-2020-contextualized} demonstrate a strong \href{https://aclweb.org/aclwiki/Question_Answering_(State_of_the_art)}{state of the art} on the AS2 task.

\section{Modeling}
\subsection{Datasets}
\label{sec:datasets}
We study popular Answer Sentence Selection datasets to evaluate the benefits of our approach for this task: Answer Sentence Natural Questions (ASNQ) \cite{garg2019tanda}, WikiQA \cite{yang-etal-2015-wikiqa}, and TrecQA \cite{voorhees99trec}.
ASNQ is a large scale QA dataset derived from Google's Natural Questions \cite{47761} dataset, with more than $\sim$84K unique questions.
The train split of this dataset is used to transfer a pre-trained transformer model to the AS2 task.
WikiQA \cite{yang-etal-2015-wikiqa} and TrecQA \cite{voorhees99trec} are widely studied benchmark datasets for Answer Sentence Selection with over $\sim$1.1K and $\sim$1.2K unique questions respectively. 
We utilize the \emph{clean} versions of both WikiQA and TrecQA, as well as the \emph{TRAIN-ALL} split of TrecQA for fine-tuning.
All of these datasets are available under dataset specific licenses that permit their use and distribution for academic purposes.

We additionally evaluate the benefits of our approach in an industry setting using the binary answer classification task using AlexaQA.
AlexaQA is a proprietary benchmark dataset that contains $\sim$107K unique questions obtained from de-identified samples of Amazon Alexa QA traffic with correct/incorrect labels assigned by expert annotators.
Detailed statistics of each dataset by split are shown in Table \ref{table:datasets} in Appendix \ref{appendix:datasets}.

\subsection{Dataset Preprocessing}
\label{sec:PP}
We implement a novel data enrichment pipeline that use an ElasticSearch index of 20.7M item and relation labels obtained by popularity-based filtering of Alexa's KB.
Our pipeline tags KB entries in input text by aggregating the results of three queries on our index.
For each word $w$ in the set of words $W=\{w_1, \dots, w_n\}$ in the input text, we tag $w_i$ as a KB entry if:
\begin{itemize}
\item{$w_i$ is an \emph{exact} match for a label in the index}
\item{$w_i$ is \emph{contained} by a label in the index}
\item{$w_i$ and $w_i+1$ is a \emph{quorum} match for a label in the index}
\end{itemize}
Consecutive labels matching the same entry are assumed joined together, matches are sorted for relevance, and the top result is selected as the KB entry.
KB meta-data for entries is derived from selected KB properties, such as the \emph{collection} property that indicates classifications such as \emph{celebrity}, \emph{book}, or \emph{album}.

\subsection{Incorporating KB-derived Context for Transformer Training}
Metadata for each entry tagged by the preprocessing pipeline (Section \ref{sec:PP}) is resolved to a textual representation using corresponding KB labels. 
An example of the JSON produced from this resolution is shown below: 
\begin{lstlisting}[language=json,firstnumber=1]
{
  "text": "David Furnish is 57 years old.",
  "kb_tags": [{
        "kb_id": "e-478772",
        "popularity": 0.981,
        "candidate_title": "David Furnish",
        "candidate_aliases": "David James Furnish, Elton John's husband"
        "collection": "celebrity",
        "relations": "married_to, years_old, birth_date, ... ",
      }]}
\end{lstlisting}

Inspired by other studies \cite{oguz2020unified, agarwal2021knowledge} that verbalize structured data for use in language models, we insert the textual representation of KB context directly into model input.
This approach may distract the model from attending to the QA pair itself if too much context is added and we thus employ two strategies to prevent this.
First, we limit metadata to the \emph{collection} property, whose values include common categories such as "celebrity", "quantity", and "generic drug form". \footnote{initial experimentation using metadata derived from the \emph{popularity}, \emph{aliases}, and \emph{relations} suggested that the \emph{collection} property was the most effective.}.
The \emph{collection} property in our KB has many analogous properties in other KBs, for example, the \emph{instance of} relation in Wikidata \cite{fardasarbas2019wikidata}.

Second, we employ a filter that constrains the number of entries from which KB context is added.
The \emph{intersection} filter exploits the intuitive hypothesis that correct QA pairs will contain the same KB entries, adding context only if the same entry is tagged in both the question and the answer.
For example, this filter adds context for entry \emph{David Furnish} from the QA pair: \emph{\textbf{Q:} how old is Elton John's husband; \textbf{A:} David Furnish is 57 years old} because the question contains \emph{Elton John's husband}, an alias for "David Furnish" in our KB, and the answer contains  \emph{David Furnish}. 
The \emph{intersection} filter excluded context for entries like \emph{57} and \emph{husband}, even though entries for both exist in our KB.
We additionally study the \emph{1-best} filter, which selects the KB entry from the answer with the highest \emph{popularity} in our KB as a more lenient alternative.
Two strategies of concatenating context to question/answer text are also explored: \emph{append} and \emph{prepend}; in both cases, the model's special separator token \footnote{We tried other separator tokens, including "\#", ":", and " ", and found the special separator performs marginally better} is used to separate the context from question/answer text. 

An example of the resulting sequences are shown below:
\begin{itemize}
	\item{\textbf{Append}: how old is elton john's husband <\textbackslash s> john furnish is 57 years old. he was born on october 25, 1962 <\textbackslash s> \emph{celebrity} <\textbackslash s> \emph{celebrity} }
	\item{\textbf{Prepend}\label{prepend}: <\textbackslash s> \emph{celebrity} <\textbackslash s> \emph{celebrity} <\textbackslash s> how old is elton john's husband <\textbackslash s> john furnish is 57 years old. he was born on october 25, 1962 }
\end{itemize}

\subsection{Model Architecture}
\label{sec:architecture4as2}

\begin{figure}[h]
  \centering
  \includegraphics[width=1\linewidth]{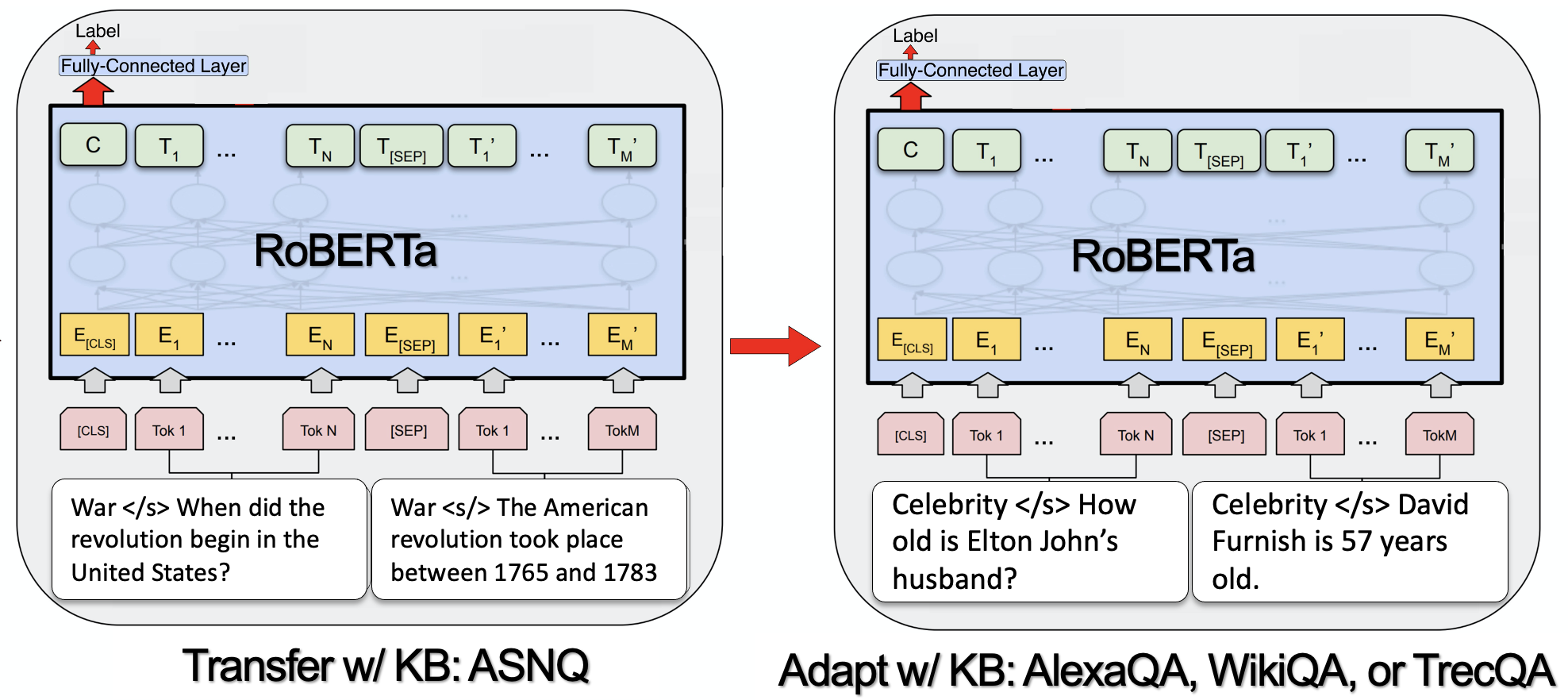}
  \caption{The Transfer-and-Adapt architecture using our approach} 
  \label{fig:tanda}
\end{figure}

Our approach builds upon the Transfer-and-Adapt (TANDA) \cite{garg2019tanda} architecture, the state of the art approach for answer sentence selection, by leveraging KB-derived context to address deficiencies observed in transformer knowledge utilization for this task.
As illustrated in figure Figure \ref{fig:tanda}, we transfer a pre-trained RoBERTa-base model \cite{liu2019roberta} to the answer sentence selection task by fine-tuning on ASNQ and adapt the transferred model via further fine-tuning on our target dataset, either WikiQA, TrecQA, or AlexaQA.  
Training incorporates KB-derived context in both transfer and adapt steps, as discussed below. 
During inference, we optionally remove KB context so as to evaluate our approach as a drop-in replacement for existing transformer-based AS2 systems. 

In order to isolate the benefits of our approach, we reuse the same optimizer, hyper-parameters, and early stopping strategy described in \cite{garg2019tanda} and only alter the sequence length, increasing from 128 to 256 to accomodate additional context.
Experiments on ASNQ, WikiQA, and TrecQA use AWS EC2 p3dn.24xlarge hosts, and those on AlexaQA use AWS SageMaker ml.p3.16xlarge notebook instances.

\section{Results}
Performance of KB augmented transformer models for standard fine-tuning (FT) on ASNQ is shown in Table \ref{table:asnq-results}. 
Transfer-and-Adapt performance with KB augmentation is reported for WikiQA, TrecQA, and AlexaQA in Tables  \ref{table:wiki-results}, \ref{table:trec-results} and \ref{table:alexa-results} respectively. 
We indicate the datasets used in Transfer-and-Adapt setting using two arguments, \textit{transfer dataset} $\rightarrow$ \textit{adapt dataset} with numerics in parentheses indicate training epochs.
Baseline models - i.e. the RoBERTa base TANDA state-of-the-art set by \cite{garg2019tanda} - are indicated by \textbf{\**} and lack the -KB suffix.

We additionally evaluate a setting in which KB context is omitted at inference time to explore the ability of our approach to modulate transformer knowledge utilization. 
Results for this setting are reported for each dataset and are indicated by the value of the \emph{Incl. KB at Inference} column. 

The results in the tables below demonstrate that:
\begin{itemize}
\item KB context improves fine-tuning performance on ASNQ, increasing the p@1, MRR and MAP by 2.9\%, 3.0\%, and 2.9\% after 9 epochs.
\item Training with KB context improves on the strong performance set by the state of the art TANDA approach on widely studied benchmarks, increasing the p@1, MRR and MAP by 2\%, 1.3\%, and 1.1\% and 4.4\%, 0.9\%, and 2.4\% on WikiQA and TrecQA respectively. 
\item The benefits of KB context generalize to our industry setting, increasing the F1 and MAP by 2.3\% and 2.0\% over the TANDA state of the art, RoBERTa ASNQ(9)$\rightarrow$AlexaQA(1), and by .4\% and 2.3\% over the more challenging baseline, RoBERTa ASNQ(1) $\rightarrow$ AlexaQA(1).
\item Models trained with our approach continue to outperform the TANDA state of the art even when KB context is omitted at inference time; in other words, the benefits of KB context are primarily realized during model training. 
\end{itemize}

\begin{table}[H]
\caption{Performance of KB-augmented fine-tuned (FT) transformer models on ASNQ}
\label{table:asnq-results}
\centering
\small
\begin{tabular}{@{}llllll@{}}
\toprule
\textbf{Model}        & \textbf{KB Approach}  & \textbf{Incl. KB at Inference} & \textbf{p@1}  & \textbf{MAP}  & \textbf{MRR}  \\
RoBERTa FT ASNQ(9)\textbf{\**}    & --                    & No                             & .599          & .672          & .716          \\
RoBERTa FT ASNQ-KB(9) & Append, Intersection  & Yes                            & .627          & .696          & .737          \\
RoBERTa FT ASNQ-KB(9) & Prepend, Intersection & Yes                            & .627          & \textbf{.702} & \textbf{.745} \\
RoBERTa FT ASNQ-KB(9) & Prepend, 1 best       & Yes                            & .616          & .694          & .736          \\
\midrule
RoBERTa FT ASNQ-KB(9) & Append, Intersection  & No                             & \textbf{.628} & .692          & .736          \\
RoBERTa FT ASNQ-KB(9) & Prepend, Intersection & No                             & .621          & \textbf{.696} & \textbf{.739} \\
RoBERTa FT ASNQ-KB(9) & Prepend, 1 best       & No                             & .617          & .693          & .735      \\
\bottomrule
\end{tabular}
\end{table}

\begin{table}[h!]
\caption{Performance of KB-augmented fine-tuned (FT) transformer models on WikiQA}
\label{table:wiki-results}
\centering
\small
\begin{tabular}{@{}llllll@{}}
\toprule
\textbf{Model}                          & \textbf{KB Approach}  & \textbf{Incl. KB at Inference} & \textbf{p@1}  & \textbf{MAP}  & \textbf{MRR}  \\
RoBERTa ASNQ(9) $\rightarrow$ WikiQA(9)\textbf{\**} & --                    & No                            & .827          & .890          & .901          \\
RoBERTa ASNQ-KB(9) $\rightarrow$ WikiQA-KB(9) & Append, Intersection  & Yes                           & .835          & .891          & .903          \\
RoBERTa ASNQ-KB(9) $\rightarrow$ WikiQA-KB(9) & Prepend, Intersection & Yes                           & \textbf{.847} & \textbf{.903} & \textbf{.913} \\
RoBERTa ASNQ-KB(9) $\rightarrow$ WikiQA-KB(9) & Prepend, 1-best       & Yes                           & .835          & .885          & .898          \\
\midrule
RoBERTa ASNQ-KB(9) $\rightarrow$ WikiQA-KB(9) & Append, Intersection  & No                            & .835          & .892          & .902          \\
RoBERTa ASNQ-KB(9) $\rightarrow$ WikiQA-KB(9) & Prepend, Intersection & No                            & \textbf{.843} & \textbf{.895} & \textbf{.907} \\
RoBERTa ASNQ-KB(9) $\rightarrow$ WikiQA-KB(9) & Prepend, 1-best       & No                            & .839          & .887          & .900         \\
\bottomrule
\end{tabular}
\end{table}

\begin{table}[h!]
\caption{Performance of KB-augmented fine-tuned (FT) transformer models on TrecQA}
\label{table:trec-results}
\centering
\small
\begin{tabular}{@{}llllll@{}}
\toprule
\textbf{Model}                                & \textbf{KB Approach}  & \textbf{Incl. KB at Inference} & \textbf{p@1}  & \textbf{MAP}  & \textbf{MRR}  \\
RoBERTa ASNQ(9) $\rightarrow$ TrecQA(9)\textbf{\**}       & --                    & No                             & .897         & .906          & .942         \\
RoBERTa ASNQ-KB(9) $\rightarrow$ TrecQA-KB(9) & Append, Intersection  & Yes                            & .911          & .901          & .952          \\
RoBERTa ASNQ-KB(9) $\rightarrow$ TrecQA-KB(9) & Prepend, Intersection & Yes                            & \textbf{.926} & \textbf{.914} & \textbf{.960} \\
RoBERTa ASNQ-KB(9) $\rightarrow$ TrecQA-KB(9) & Prepend, 1-best       & Yes                            & .897          & .900          & .944          \\
\midrule
RoBERTa ASNQ-KB(9) $\rightarrow$ TrecQA-KB(9) & Append, Intersection  & No                             & \textbf{.941} & \textbf{.915} & \textbf{.966} \\
RoBERTa ASNQ-KB(9) $\rightarrow$ TrecQA-KB(9) & Prepend, Intersection & No                             & .911          & .901          & .955          \\
RoBERTa ASNQ-KB(9) $\rightarrow$ TrecQA-KB(9) & Prepend, 1-best       & No                             & .926          & .905          & .959         \\
\bottomrule
\end{tabular}
\end{table}

\begin{table}[h!]
\caption{Performance of KB-augmented fine-tuned (FT) transformer models on AlexaQA. Models transferred for only (1) epoch are shown, since our experiments indicate that further epochs of transfer to ASNQ conveyed marginal benefits for AlexaQA.}
\label{table:alexa-results}
\centering
\small
\begin{tabular}{@{}lllll@{}}
\toprule
\textbf{Model}                                & \textbf{KB Approach}  & \textbf{Incl. KB at Inference} & \textbf{F1}   & \textbf{MAP}  \\
\midrule
RoBERTa ASNQ(1) $\rightarrow$ AlexaQA(1)      & --                    & No                            & .848          & .839          \\
RoBERTa ASNQ(9) $\rightarrow$ AlexaQA(1)\textbf{\**}      & --                    & No                            & .829          & .842          \\
RoBERTa ASNQ-KB(1) $\rightarrow$AlexaQA-KB(1) & Append, Intersection  & Yes                           & \textbf{.852} & .860          \\
RoBERTa ASNQ-KB(1)$\rightarrow$AlexaQA-KB(1)  & Prepend, Intersection & Yes                           & .850          & \textbf{.862} \\
RoBERTa ASNQ-KB(1)$\rightarrow$AlexaQA-KB(1)  & Prepend, 1-best       & Yes                           & .850          & .858          \\
\midrule
RoBERTa ASNQ-KB(1)-$\rightarrow$AlexaQA-KB(1) & Append, Intersection  & No                           & \textbf{.851} & .859          \\
RoBERTa ASNQ-KB(1)$\rightarrow$AlexaQA-KB(1)  & Prepend, Intersection & No                            & .850 & \textbf{.861} \\
RoBERTa ASNQ-KB(1)$\rightarrow$AlexaQA-KB(1)  & Prepend, 1-best       & No                            & .849          & .857         \\
\bottomrule
\end{tabular}
\end{table}

\section{Discussion}
\subsection{Comparing Context Generation Strategies} 
Results reported in Tables \ref{table:asnq-results}, \ref{table:wiki-results}, \ref{table:trec-results} and \ref{table:alexa-results} all demonstrate that our approach outperforms the state of the art approach, even in the more challenging setting where KB context is omitted at inference time.
We explain the robustness of our models to the omission of KB context in light of the proportion of each dataset that our approach impacts.
The \emph{intersection} filter adds KB context to only 3.38\% of the ASNQ dataset, 5.27\% of TrecQA, and 8.33\% of WikiQA while the \emph{1 best} filter adds context for 31.17\% for ASNQ, 51.1\% for TrecQA, 40.79\% for WikiQA.
We hypothesize that the large number of training examples seen without context allows the model to leverage context as a for weak supervision that encourages knowledge utilization and elaborate further in subsection \ref{sec:kb-context} below. 

% TODO AMLC feedback here? 
These results show that the more intuitive \emph{intersection} filter performs better than the \emph{1 best} filter for both concatenation strategies, despite impacting between significantly less of each dataset.
We conclude that the explicit conceptual alignment provided by the \emph{intersection} conveys additional benefits beyond the addition of conceptual keywords provided by the \emph{1 best} filter.
The \emph{prepend} strategy outperforms the \emph {append} strategy on all datasets other than TrecQA, a deviation that we attribute to the small size of the TrecQA test set.
We explicate these findings in light of the positional invariance the \emph{prepend} strategy - that is, prepend always adds context in the same position in the sequence, whereas \emph{append} does not.
As a result, \emph{prepend} models appear better able to attend to context and outperform their \emph{append} counterparts, even though \emph{prepend} models suffer more when context is omitted at inference.

\subsection{Impact of KB Context} 
\label{sec:kb-context}
We leverage the three illustrative examples presented in Table \ref{table:hard-qa-pairs} to probe the impact of our KB context and its potential to address the previously studied \cite{jiang2020know, talmor2020olmpics} deficiencies of transformers as implicit KBs.
Models trained with our approach classify each of these examples correctly, even when KB is omitted at inference, indicating that they may be able to exploit our context to refine their utilization of encoded knowledge.
In order to identify the mechanism behind these benefits, we compare the attention of TANDA with that of our best model, \emph{prepend, intersection}, using box plots of attention intensity and bar plots of activate head counts per layer in Appendix \ref{appendix:attention}.

\textbf{Example 1} requires the model to leverage encoded knowledge in order to make the connection between \emph{"husband"} and \emph{"David Furnish"} necessary to recognize that the phrase \emph{"is 57 years old"} answers the question phrase \emph{"how old"}.
Figure \ref{fig:elton-john-viz} presents model attention weights between tokens \emph{"how"} and \emph{"57"} and between \emph{"husband"} and \emph{"David"}, where it can be seen that our approach significantly improves both the quantity of heads attending to these keywords and the intensity of this attention.
It is likely that model pre-training has encoded this knowledge, given that the second sentence on David Furnish's Wikipedia page reads: \emph{"He is married to English musician Sir Elton John"}. 
Unsurprisingly, changing the question or the answer text to remove this relation - to either \emph{"How old is David Furnish"} or \emph{"Elton John's husband David Furnish is 57 years old"} - produces the correct answer from the TANDA model.

\textbf{Example 2} probes transformer ability to robustly recognize that \emph{"one-humped"} and \emph{"two-humped"} are values for the quantity sought by \emph{"how many"} and are related to the subject \emph{"Camel"}.
We hypothesize that the KB context \emph{"animal"} added for similar entities during training increases attention on \emph{"camel"} tokens and their modifiers, \emph{"one-humped"} and \emph{"two-humped"} in this case. 
Figure \ref{fig:camel-viz} compares model attention weights of tokens \emph{"many"} and \emph{"Camel"} with the values \emph{"one"} and \emph{"two"} and again demonstrate that our approach significantly increases the intensity of model attention between these terms.
Changing the answer to use common numeric values \emph{"the Dromedary Camel has 1 hump...and the Bactrian Camel has 2 humps"} is sufficient for the TANDA model to select the correct answer.

\textbf{Example 3} illustrates whether the model is able to connect the adverbial phrase \emph{"some legal uses"} in the question with the phrase \emph{approved...for the treatment of...} in the correct answer.
Interestingly, the KB context added for \emph{"meth"} and entities like it is \emph{"generic drug"}, which we hypothesize may encourage attention to relevant terms like \emph{"treatment"} that are not commonly used in context of the subject \emph{"meth"}.
Figure \ref{fig:meth-viz} shows the weights connecting \emph{"treatment"} with \emph{"uses"} and \emph{"meth"} and further demonstrates the impact of our approach on model attention.
We conclude that in some cases, the context itself may provide relevant information that helps the model more effectively utilize uncommon knowledge, like that meth may be used as a medical treatment. 

\section{Conclusion}
In this paper, we presented a data-programming approach that enriches transformer training data with KB-derived context, and demonstrate that it beats state of the art approach on several challenging knowledge-intensive question-answering benchmarks such as ASNQ, WikiQA, TrecQA, and Alexa QA. 
Our findings indicate that our approach addresses some deficiencies of transformer knowledge utilization that negatively impact AS2 performance.
We probed the mechanism of our approach with challenging examples that highlight the potential ways in which our KB context may allow transformers to better utilize encoded knowledge.
Our method is simple, efficient and task-agnostic, and training benefits remain even when KB context is omitted at inference time. 
We believe that our approach provides a way to rapidly integrate the benefits of KBs within the deployed inference pipelines utilized in many virtual-assistant workflows.

While we improve on the state of the art approach in AS2, we do acknowledge that our approach may face limitations of its own. 
While our approach is efficient in that it not require significant pre-training, unlike KB based approaches like KELM, KnowBERT, and ERNIE as well as retrieval oriented approaches like REALM and RAG, it is inefficient in that it likely does not leverage the full richness of our KB.
This has the negative consequence that our approach still requires significant task-specific training and thus consumes significant GPU hours and the natural resources used to power them.
Further work beyond the data-programming approach that we propose in the direction of more effective transformer architectures that enhance knowledge utilization can lessen this impact and provide models capable of more completely disentangling knowledge and language acquisition. 

\bibliographystyle{acm}
\bibliography{custom}

\appendix
\section{Dataset statistics}
\label{appendix:datasets}
The table below shows the distribution of the datasets studied in this by each split, such as \emph{train}, \emph{dev}, and \emph{test}. 
These demonstrate that our pipeline is able to tag at least one KB entry in each input QA pair, indicating that our simple tagging method is effective at producing KB context. 

\begin{table}[H]
\centering
\small
\caption{Dataset Statistics and KB Tag Rate by Split}
\label{table:datasets}
\begin{tabular}{rrrrr}
\toprule
\multicolumn{1}{c}{\textbf{Dataset}} & \multicolumn{1}{c}{\textbf{\#QA pairs}} & \multicolumn{1}{c}{\textbf{\% w/o KB}} & \multicolumn{1}{c}{\textbf{\#Correct w/ KB}} & \multicolumn{1}{c}{\textbf{\#Incorrect w/ KB}} \\
\midrule
ASNQ Dev                             & 276,809                                 & .020\%                                 & 1,117                                        & 275,692                                        \\
ASNQ Test                            & 879,594                                 & .036\%                                 & 3,600                                        & 875,672                                        \\
ASNQ Train                           & 29,987,324                              & .027\%                                 & 120,184                                      & 29,867,166                                     \\
\midrule
WikiQA Dev                           & 1,130                                   & .000\%                                 & 140                                          & 990                                            \\
WikiQA Test                          & 2,351                                   & .000\%                                 & 293                                          & 2,507                                          \\
WikiQA Train                         & 8,672                                   & .000\%                                 & 1,040                                        & 7,632                                         \\
\midrule
TrecQA Dev                           & 1,117                                   & .000\%                                 & 205                                          & 912                                            \\
TrecQA Test                          & 1,442                                   & .000\%                                 & 248                                          & 1,194                                          \\
TrecQA Train                         & 53,417                                   & .000\%                                 & 6,403                                        & 47,011                                         \\
\midrule
AlexaQA Dev                          & 26,951                                  & .040\%                                 & 25,822                                       & 1,192                                          \\
AlexaQA Test                         & 26,965                                  & .000\%                                 & 25,796                                       & 1,169                                          \\
AlexaQA Train                        & 215,416                                 & .635\%                                 & 205,070                                      & 8,978                                          \\
\bottomrule
\end{tabular}
\end{table}

\section{Attention Weight Comparison}
\label{appendix:attention}
In the graphs below, we illustrate the impact of our approach on model attention for the challenging AS2 examples presented in Table \ref{table:hard-qa-pairs}.
We do not add KB context at inference for any of these examples, opting to visualize the impact of our approach in the more challenging "omit KB" setting. 
We leverage BertViz \cite{vig2019multiscale} to extract model attention weights and quantify model attention between meaningful keywords selected in question and answer texts. 
Box plots, shown on the left, quantify the intensity of model attention across all layers, while bar plots, shown on the right, quantify the number of heads per layer exhibiting attention weights greater than an arbitrary minimum of 0.1. 

\begin{figure}[h]
  \centering
  \includegraphics[width=.8\linewidth]{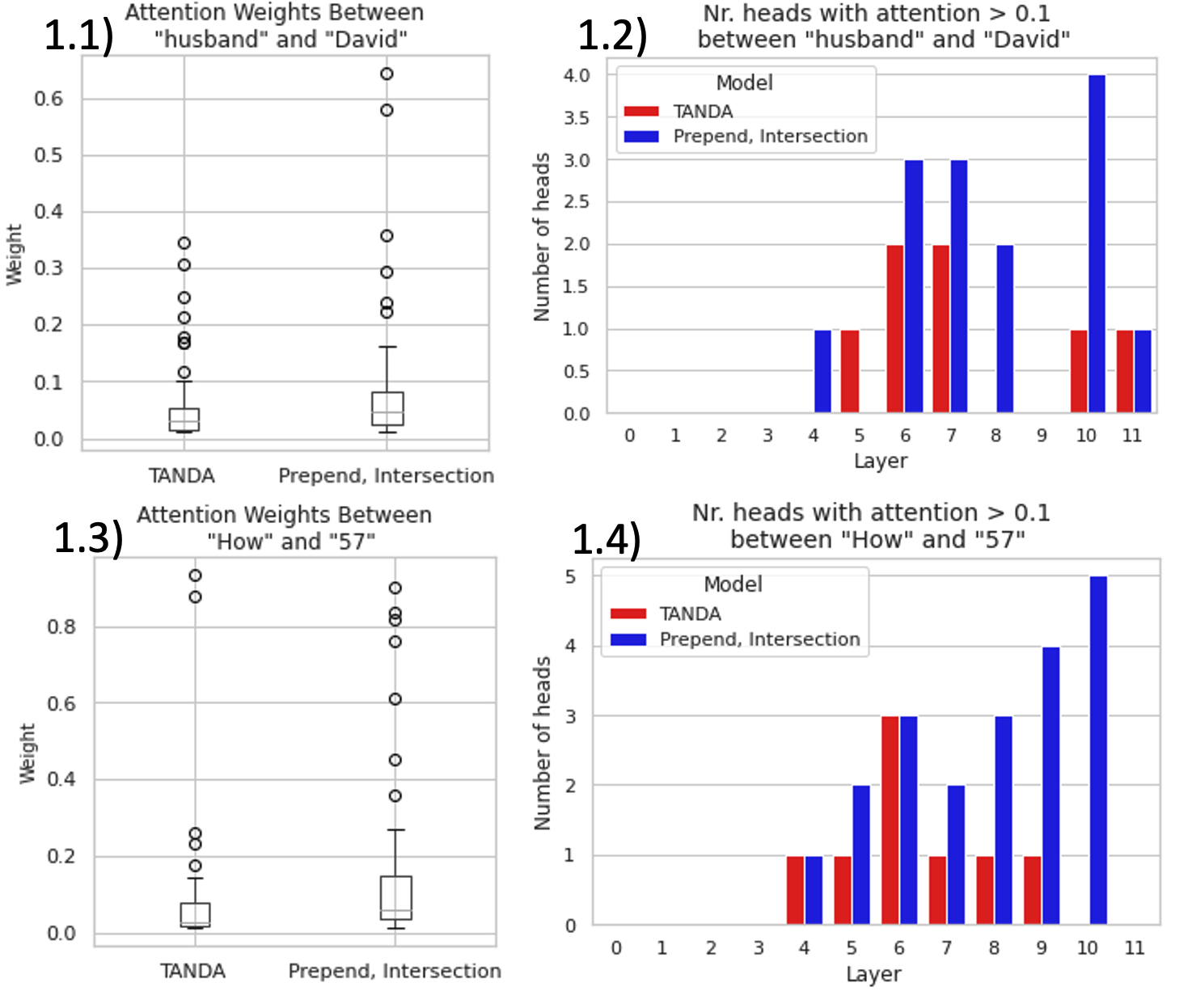}
  \caption{Attention comparison for the correct QA pair \emph{\textbf{Q:} How old is Elton John's husband \textbf{A:} David Furnish is 57 years old. He was born on October 25, 1962}}. 
  \label{fig:elton-john-viz}
\end{figure}

\begin{figure}[h]
  \centering
  \includegraphics[width=.8\linewidth]{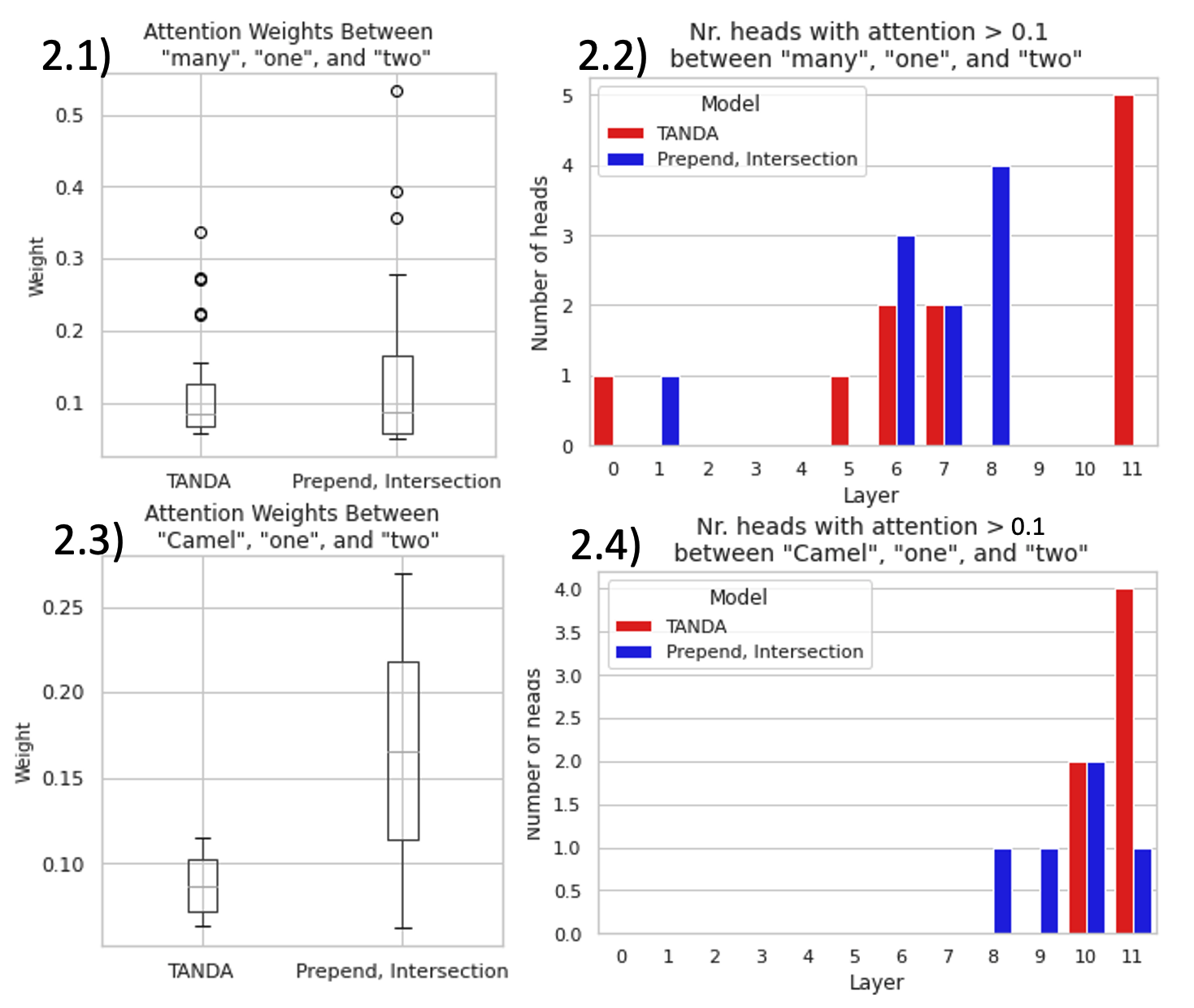}
  \caption{Attention comparison for the correct QA pair \emph{\textbf{Q:} How many humps on a Camel? \textbf{A:} The two surviving species of camel are the dromedary, or one-humped camel, which is native to the Middle East and the Horn of Africa; and the Bactrian, or two-humped camel, which inhabits Central Asia.}
  }
  \label{fig:camel-viz}
\end{figure}

\begin{figure}[h]
  \centering
  \includegraphics[width=.8\linewidth]{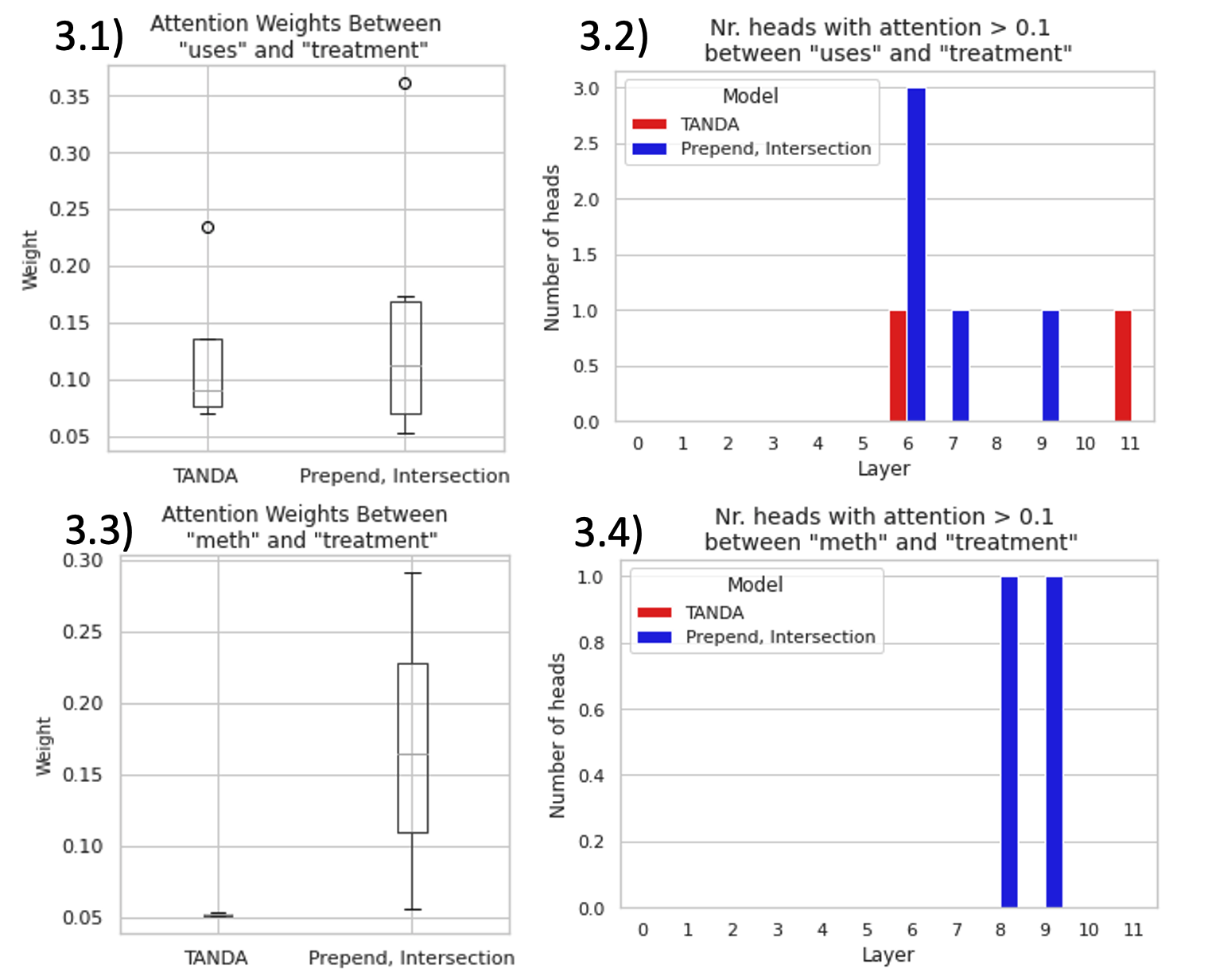}
  \caption{Attention comparison for the correct QA pair \emph{\textbf{Q:} What some legal uses of meth? \textbf{A:} Although rarely prescribed, methamphetamine hydrochloride is approved by the U.S. Food and Drug Administration (FDA) for the treatment of attention deficit hyperactivity disorder and obesity under the trade name Desoxyn.}
  }
  \label{fig:meth-viz}
\end{figure}

\end{document}